\documentclass[letterpaper]{article}

\usepackage{natbib,alifeconf}  %% The order is important

\usepackage{amsmath}

\usepackage{url,hyperref,cleveref}
\usepackage{booktabs}

\usepackage{graphicx}
\usepackage{float}
\usepackage{subcaption}

\title{Flow-Lenia.png: Evolving Multi-Scale Complexity by Means of Compression}

\author{
    Tadashi Adachi$^{1*}$,
    Solvi Arnold$^{2}$,
    Takafumi Mochizuki$^{1}$ \and
    Kimitoshi Yamazaki$^{2}$
    \mbox{}\\
    $^1$EPSON AVASYS, Japan \\
    $^2$Shinshu University, Japan \\
    $^*$Adachi.Tadashi2@exc.epson.co.jp
} % email of corresponding author

\begin{document}

\maketitle

\begin{abstract}
    We propose a fitness measure quantifying multi-scale complexity for cellular automaton states, using  compressibility as a proxy for complexity. We explore the complexity range accessible to Flow Lenia, and demonstrate evolution of patterns of specific complexity.
\end{abstract}

\section{Introduction}

Flow Lenia \citep{plantec2023flowlenia}, an extension of Lenia \citep{Chan_2019} incorporating mass conservation constraints, facilitates generation of stable spatially localized patterns and discovery of intriguing artificial organisms. The generated patterns reveal a diverse array of complex shapes, ranging from minute cell clusters to intricate patterns composed of multiple organs.
When using such systems to study open-ended evolution (OEE) and the emergence of complex life, it is useful to understand the range of complexity that a given system configuration can produce, and where within that complexity landscape patterns of interest reside.

In this study, we propose a fitness function that quantifies complexity as multi-scale compressibility, and present experimental results where this fitness function is used to let Flow Lenia evolve towards 1) specific complexity targets, and 2) the extrema of its complexity domain (under a given hyperparameter configuration).

\section{Related Work}

In Lenia, indicators like angular speed and symmetry are employed for species classification. Subsequent work \citep{Chan2023TowardsLS} pursued open-ended evolution (OEE) in Lenia, and noted the necessity of quantitative assessments as one of several factors that can facilitate successful simulations to achieve OEE.

\begin{figure}[ht!]
    \centering
    \includegraphics[width=73mm]{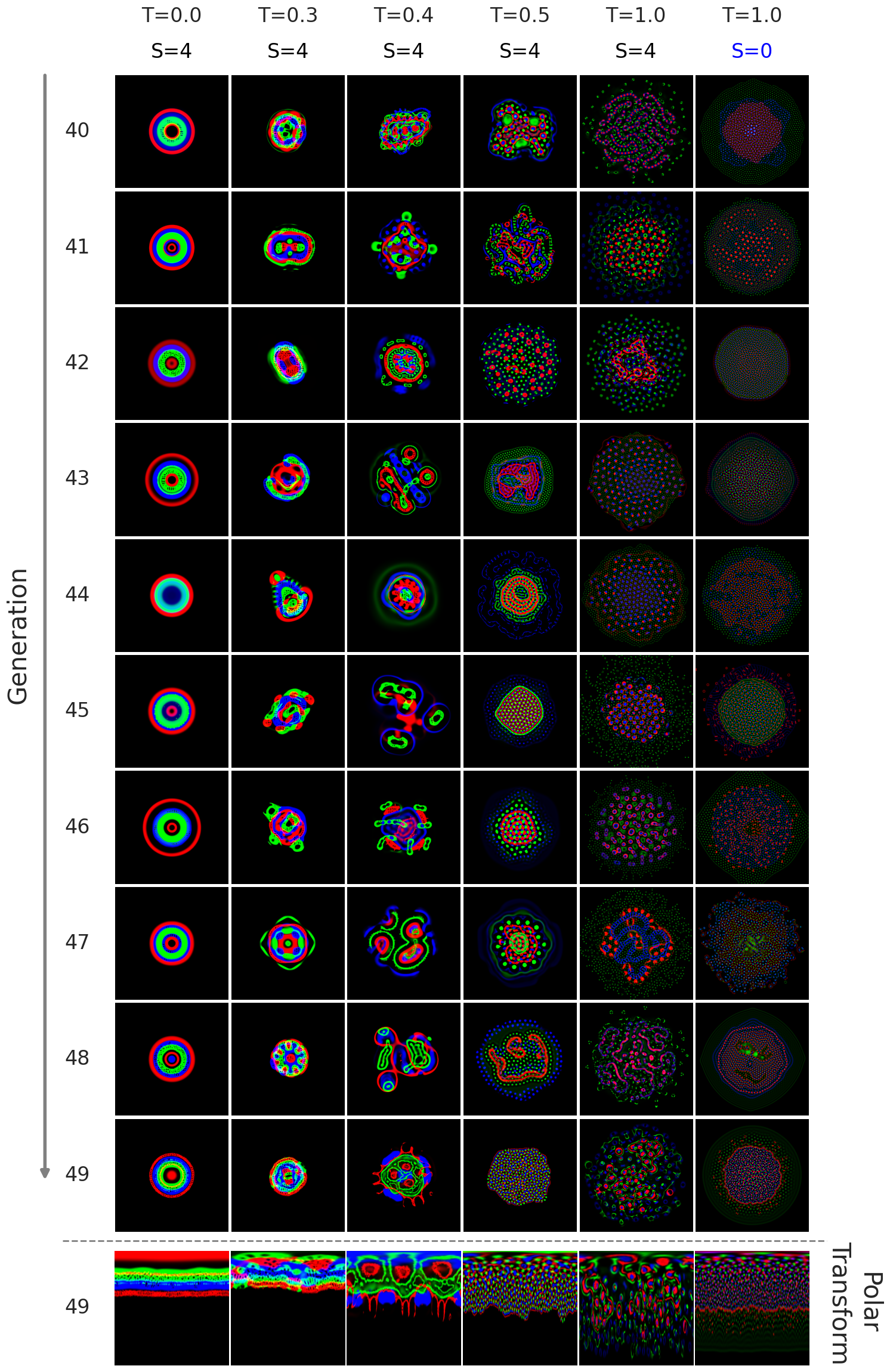}
    \caption{\small Fittest individuals of the last 10 generations of evolutionary trials with various settings for T and S. The bottom row shows examples of polar-transformed states for each setting.}
    \label{fig:multiscale}
\end{figure}

\begin{figure*}[ht]
    \centering
    \includegraphics[width=\linewidth]{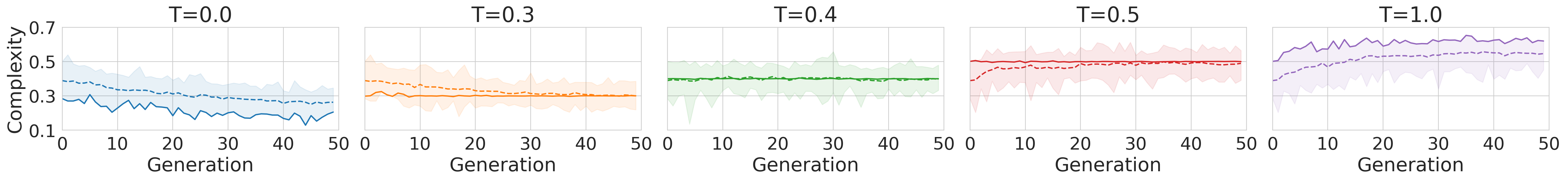}
    \includegraphics[width=\linewidth]{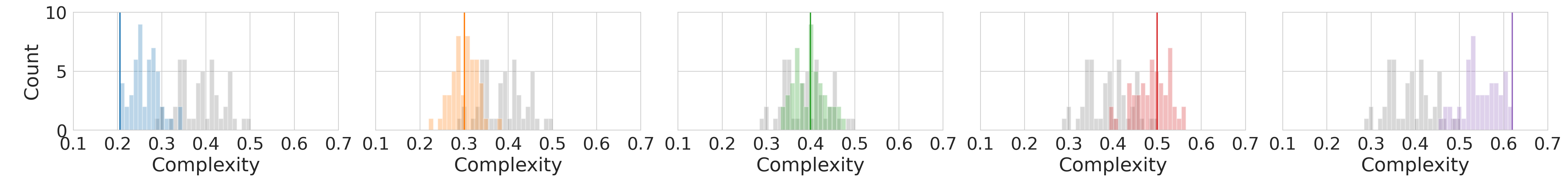}
    \caption{\small Top row: Complexity Evolution Trends. Bold line: \textit{Best}; Dashed line: \textit{Mean}; Light color: \textit{Min-Max range}. Bottom row: Complexity distribution in the final population. Gray: \textit{Initial generation}; Color: \textit{Final generation}; Bold line: \textit{Best}.}
    \label{fig:overall_figure}
\end{figure*}

\section{Method}

Kolmogorov complexity \citep{solomonoff1960} \citep{kolmogorov65} defines the complexity of an object in terms of the size of its smallest possible representation. The higher an object's compressibility, the lower its complexity. Exact calculation of Kolmogorov complexity is infeasible in general. However, here we aim to quantify complexity of Flow Lenia states, which are image-like objects. For images, highly effective compression methods are widely available. Consistent with Kolmogorov complexity, uniform images are highly compressible, while noise images are not compressible at all. Accordingly, we use the ratio between the compressed and uncompressed representation of a state as  proxy of state complexity. We can then propose the following fitness function \ref{Fitnessfunction} to quantify how closely a given state matches a given multi-scale complexity target:

\begin{equation}
\label{Fitnessfunction}
\text{Fitness}(x^i_g) = \frac{1}{S+1} \sum_{s=0}^{S} \left| C(x^i_g, s) - T \right|
\end{equation}

where \( x^i_g \) is individual \( i \) in generation \( g \), \( C(x, s) \) is the inverse compression ratio of \( x \) after resampling to \( 2^{-s} \) times its original resolution, T is the complexity target, and \( S \) controls the number of scales considered. In preliminary experiments, we observed that due to the use of radial diffusion kernels, Flow Lenia often produces radial patterns. Rasterizing these patterns into a square cell grid obscures their radial symmetry, causing image compression rates to overestimate the pattern's actual complexity. To counteract this effect, we re-rasterize state representations in a polar coordinate system before calculating compression rates.

\section{Experiments}

We evolve the parameters controlling Flow Lenia's state update logic using a Genetic Algorithm. Hyperparameter settings and search space definition follow \citep{plantec2023flowlenia}, using 3 channels, 12 kernels, and 8 evolvable parameter types (R, r, h, a, b, w, $\mu$, $\sigma$). The world size is set to 256x256. We randomly initialize a population of 50 individuals, and evolve it for 50 generations, using rank-based selection, uniform crossover, and point mutation at a 5\% mutation probability. To obtain the fitness value for a given parameter set, we run Flow Lenia with it for 2000 state updates, and apply our fitness function to the final resulting state. Compression ratios are obtained using the lossless PNG compression algorithm \citep{crocker1995png}.

We ran experiments with complexity target T set to 0.0, 0.3, 0.4, 0.5, and 1.0. T=0.0 and T=1.0 correspond to the extrema of our complexity measure. These experiments let us identify the minimal and maximal complexity that this configuration of Flow Lenia can produce. Targets 0.3 through 0.5 explore the region between the observed extrema. The multi-scale parameter S is set to 4 by default. As its effects are most apparent in combination with higher T values, we run an additional experiment with T=1.0 and S=0 (single-scale complexity assessment).

\section{Results}

Figure \ref{fig:multiscale} shows the fittest individual of the final 10 generations in each experiment. As T increases, evolved patterns become more intricate, consistent with human perception of complexity. For T=1.0, we show patterns obtained with single- and multi-scale complexity targets. With S=0, we obtain fine granular patterns. We can understand this as Flow Lenia's best approximation of pure noise. These patterns maximize complexity at a fine-grained level, but assessed at larger scales they are uniform rather than noise-like. When we "zoom out" (i.e. downsample), these patterns quickly become compressible (as do noise images). With S=4, larger structures form over granular backdrops, demonstrating the effectiveness of the multi-scale feature.

In Figure \ref{fig:overall_figure} (top row), we observe that mean complexity gradually approaches the target T over successive generations. The bottom row of Figure \ref{fig:overall_figure} shows the complexity distribution of initial and final populations. The complexity distribution for the initial generation shows that randomly sampled individuals center around 0.4. Results for T=0.0 and T=1.0 together identify the complexity range accessible to this system configuration, finding lower and upper bounds at about 0.21 and 0.62 respectively. For T=0.3, T=0.4, and T=0.5 the complexity distribution is seen to center on T. This demonstrates that within the bounds established above, our fitness function allows us to evolve patterns matching specific complexity objectives.

\section{Discussion and Future Work}

In the present work, we relied on crossover and mutation to maintain diversity. A future direction is to combine complexity targets with explicit selection for novelty, to effectively explore pattern diversity within specific complexity regions. Another direction of interest is to track how complexity develops under selection for specific abilities (e.g. locomotion), to explore how various selection pressures foster or obstruct evolution towards higher complexity.

% \newpage

\footnotesize
\bibliographystyle{apalike}
\bibliography{references}

\end{document}